\documentclass{article}

 \usepackage[final,nonatbib]{nips_2016}

\usepackage[utf8]{inputenc} 
\usepackage[T1]{fontenc}    
\usepackage{hyperref}       
\usepackage{url}            
\usepackage{booktabs}       
\usepackage{amsfonts}       
\usepackage{amsmath}
\usepackage{graphicx}
\usepackage{amssymb}
\usepackage{nicefrac}       
\usepackage{microtype}      
\usepackage{color}

\newcommand{\x}{\boldsymbol{x}}

\title{Interpreting the Predictions of Complex ML Models by Layer-wise Relevance Propagation}

\author{
  Wojciech Samek\\
Fraunhofer HHI, 10587 Berlin, Germany\\
  \texttt{\small wojciech.samek@hhi.fraunhofer.de}
  \And
  Gr\'egoire Montavon\\
TU Berlin, 10587 Berlin, Germany\\
  \texttt{\small gregoire.montavon@tu-berlin.de}
  \AND
  Alexander Binder\\
  SUTD, 487372 Singapore\\
  \texttt{\small alexander\_binder@sutd.edu.sg}
  \And
  Sebastian Lapuschkin\\
Fraunhofer HHI, 10587 Berlin, Germany\\
  \texttt{\small sebastian.lapuschkin@hhi.fraunhofer.de}
  \AND
  Klaus-Robert M\"uller\\
TU Berlin, 10587 Berlin, Germany\\
  \texttt{\small klaus-robert.mueller@tu-berlin.de}
}

\begin{document}

\maketitle

\begin{abstract}

Complex nonlinear models such as deep neural network (DNNs) have become an important tool for image classification, speech recognition, natural language processing, and many other fields of application. These models however lack transparency due to their complex nonlinear structure and to the complex data distributions to which they typically apply. As a result, it is difficult to fully characterize what makes these models reach a particular decision for a given input. This lack of transparency can be a drawback, especially in the context of sensitive applications such as medical analysis or security. In this short paper, we summarize a recent technique introduced by Bach et al.\ \cite{BacBinMonKlaMueSam15} that explains predictions by decomposing the classification decision of DNN models in terms of input variables.

\end{abstract}

\section{Explaining Predictions by Decomposition}

Nonlinear models such as neural networks are an essential component of many practical machine learning algorithms. These models have solved numerous practical problems such as visual object recognition, speech recognition, or natural language processing. Deep neural networks in particular, have been shown recently to perform extremely well on these tasks \cite{krizhevsky2012imagenet, hinton2012deep, kim2014convolutional}. A typical limitation of complex machine learning models is their tendency to predict in a black-box manner, providing little information on what aspect of the input data supports the actual prediction. The problem of explaining neural network predictions \cite{DBLP:journals/corr/SimonyanVZ13b, DBLP:conf/eccv/ZeilerF14, BacBinMonKlaMueSam15, zhang2016top, nguyen2016multifaceted, shrikumar2016not, kasneci2016licon, ribeiro2016should} has received a lot of attention recently, especially in the context of image recognition with convolutional neural networks. From the multitude of recently proposed methods for explaining predictions, we can identify \emph{decomposition approaches} \cite{DBLP:conf/aaai/PoulinESLGWFPMA06, DBLP:conf/cidm/LandeckerTBMKB13, BacBinMonKlaMueSam15, MonArXiv15}, which seek to redistribute the value of the prediction function on the input variables, such that the sum of redistributed terms corresponds to the actual function value.

Let us formalize the problem of explaining a prediction from the perspective of decomposition. Let $f:\mathbb{R}^d \to \mathbb{R}$ be the prediction function, $\x$ a data point given as input to the model and $f(\x)$ the prediction for this particular data point. In the context of image classification, the data point is formed by a set of pixels: $\x = (x_p)_p$. Explanation through the decomposition framework produces a vector of scores $(R_p)_p$ associated to each pixel, indicating how relevant pixels are to the prediction. These relevance scores are a decomposition if they satisfy the conservation equation $\sum_p R_p = f(\x)$. In practice, other criteria are necessary to define a good decomposition, for example the ratio between positive and negative scores, and the relation between the relevance score and the effect of the corresponding pixel activation on the predicted function value.

In contrast to sensitivity analysis which assigns scores based on the effect of a small or infinitesimal pixel perturbation on the function value, e.g. $(\partial f / \partial x_p)^2$, thus producing an explanation of a local variation of $f$, the decomposition approach on the other hand seeks to explain the whole function value (i.e. what made the function reach a certain value and not zero). Practically, when considering, for example, an image of the class ``scooter'', sensitivity analysis tells us which pixels in the image, if modified, makes the image more or less belong to that class. Instead, the decomposition approach explains what pixels speak in what amount for the presence of the scooter in the image. Figure~\ref{fig:scooter}~(a) shows the qualitative difference between sensitivity analysis and decomposition (with the latter computed using the LRP method described below).

The difference between decomposition and sensitivity analysis can also be illustrated from the perspective of a linear classifier $f(\x) = \sum_p x_p w_p$. A possible decomposition is given by identifying relevance scores as the terms being summed: $R_p = x_p w_p$. Instead, sensitivity analysis will return $R_p = w_p^2$ (or $R_p = |w_p|$ depending on the variant), which does not involve the actual pixel activation. If in particular $x_p$ quantifies the presence of a local feature at position $p$, it would be intuitive that the feature is considered relevant if not only the classifier reacts to it ($w_p>0$), but also if that feature is actually present ($x_p>0$). Only the decomposition approach takes these two parameters into account.

\section{Layer-wise Relevance Propagation}

Layer-wise relevance propagation (LRP) \cite{BacBinMonKlaMueSam15} is one such method, that operates by building for each neuron of a deep network a local redistribution rule, and applying these rules in a backward pass in order to produce the pixel-wise decomposition. LRP has been successfully applied to many different models and tasks beyond classification of images by convolutional neural networks. For instance, in \cite{BacBinMonKlaMueSam15} and \cite{LapCVPR16} it has been applied to Bag-of-Words models and Fisher Vector / SVM  classifiers, respectively. In \cite{ArrACL16} it was used for the identification of relevant words in text documents and in \cite{ArbGCPR16} for visualizing facial features related to age, happiness and attractivity. Also the authors of \cite{StuJNE16} use LRP for identifying relevant spatio-temporal EEG features in the context of Brain-Computer Interfacing.

Within the LRP framework, various rules have been proposed for redistributing the relevance assigned to a neuron onto its input neurons. Let $(x_i)_i$ be the neuron activations at layer $l$. Let $(R_j)_j$ be the relevance scores associated to the neurons at layer $l+1$. Let $w_{ij}$ be the weight connecting neuron $i$ to neuron $j$. The ``alpha-beta'' rule \cite{BacBinMonKlaMueSam15} redistributes relevance from layer $l+1$ to layer $l$ in the following way:
\begin{align}
  R_{i} = \sum_{j} \Big(
   \alpha\cdot \frac{(x_i w_{ij})^+}{\sum_i (x_i w_{ij})^+} -
   \beta \cdot \frac{(x_i w_{ij})^-}{\sum_i (x_i w_{ij})^-}   
  \Big) R_{j},
    \label{eq:secondrule}
\end{align}
where $()^+$ and $()^-$ denote the positive and negative parts, respectively. Layer-wise conservation of relevance ($\sum_i R_i = \sum_j R_j$) is enforced by choosing $\alpha,\beta$ such that $\alpha-\beta=1$. Choosing the parameters $\alpha=2$ and $\beta=1$ was shown to produce nice-looking and sharp heatmaps. This set of parameters works well for a wide range of neural network models, and allows to express contradicting evidence in the input image, through negative relevance scores. Example of heatmaps produced with these parameters are shown in Figure \ref{fig:scooter}. Choosing instead the parameters $\alpha=1$ and $\beta=0$ and assuming positive activations provides connections to other methods. In particular, it yield the simplified formula:
\begin{align}
  R_{i} = \sum_{j} \frac{x_i w_{ij}^+}{\sum_i x_i w_{ij}^+} R_{j}
    \label{eq:secondrule-alpha}
\end{align}
which is equivalent to the $z^+$-rule by \cite{MonArXiv15} and the redistribution rule used by \cite{zhang2016top} as part of the Excitation Backprop (EB) method\footnote{Compare Eq.~(2) in \cite{zhang2016top} and the $z^+$ rule in \cite{MonArXiv15}. Since the authors of \cite{zhang2016top} assume the response of the activation neuron to be non-negative, Eq.~(2) in \cite{zhang2016top} is also equivalent to the alpha-beta LRP rule \cite{BacBinMonKlaMueSam15} for $\alpha=1$ and $\beta=0$.}. Futhermore, \cite{MonArXiv15} showed that this particular redistribution rule results from a ``deep Taylor decomposition'' of the neural network function when the neural network is composed of ReLU neurons. The main concepts used by deep Taylor decomposition and how it leads to the propagation rule of Equation \ref{eq:secondrule-alpha} are outlined in the following.

\paragraph{Connection to Deep Taylor Decomposition} Assume that the relevance score $R_j$ is the product of the ReLU activation $x_j$ and a positive constant term $c_j$. It can then be written as:
\begin{align*}
\textstyle R_j
&= x_j c_j\\
&= \textstyle \max(0,\sum_i x_i w_{ij} + b_j) \cdot c_j\\
&= \textstyle \max(0,\sum_i x_i w_{ij} c_j + b_j c_j),
\end{align*}
where $w_{ij} c_j$ and $b_j c_j$ are the weights and bias parameters of a newly defined ``relevance neuron''. A first-order Taylor expansion of the relevance neuron at some reference point $(\widetilde x_i)_i$ allows to decompose the neuron value in terms of its input neurons. The reference point is chosen to be the intersection of the neuron equation $\sum_i x_i w_{ij} c_j + b_j c_j = \varepsilon$ with $\varepsilon$ positive and infinitesimally small, and the search line $\{(x_i)_i - t \cdot (x_i 1_{\{w_{ij} c_j>0\}})_i \colon t \in \mathbb{R}^+\}$, that is, progressively deactivating positively contributing inputs until the relevance becomes almost zero. In that case, the relevance ``flowing'' from neuron $j$ to neuron $i$ is given by the identification of the corresponding first-order term of the Taylor expansion and has a closed form solution:
$$
R_{i \leftarrow j} = \frac{\partial R_j}{\partial x_i}\Big|_{(x_i)_i = (\widetilde x_i)_i} \!\! \cdot (x_i - \widetilde x_i)
= \frac{x_i (w_{ij} c_j)^+}{\sum_i x_i (w_{ij} c_j)^+} R_j
= \frac{x_i w_{ij}^+}{\sum_i x_i w_{ij}^+} R_j.
$$
When we pool the relevance messages coming from the upper-layer neurons ($R_i = \sum_j R_{i \leftarrow j}$), we recover Eq.~(\ref{eq:secondrule-alpha}). We now show that the product structure on which the decomposition relies also holds approximately for the lower-layer relevance. This can be made visible by rewriting Eq.~(\ref{eq:secondrule-alpha}) as
$$
R_{i} = \sum_{j} \frac{x_i w_{ij}^+}{\sum_i x_i w_{ij}^+} R_{j} = x_i \cdot \underbrace{\sum_{j} \frac{w_{ij}^+  \cdot \max(0,\sum_i x_i w_{ij} + b_j) \cdot c_j}{\sum_i x_i w_{ij}^+}}_{c_i}
$$
where $c_i$ is positive and can indeed be considered as approximately constant due to its very weak dependence on $x_i$ (diluted by two nested sums). Thus, if the decomposition can be performed at a certain layer, it can also be performed at the previous layer. For more details we refer the reader to the original paper \cite{MonArXiv15}.

\section{LRP in Practice}

In this section, we briefly discuss how LRP compares to sensitivity analysis, and how LRP can be used in practice for comparing machine learning models and testing what strategy these models use to predict the data.

\begin{figure}[h]
\centering \small
\includegraphics[width=0.85\textwidth]{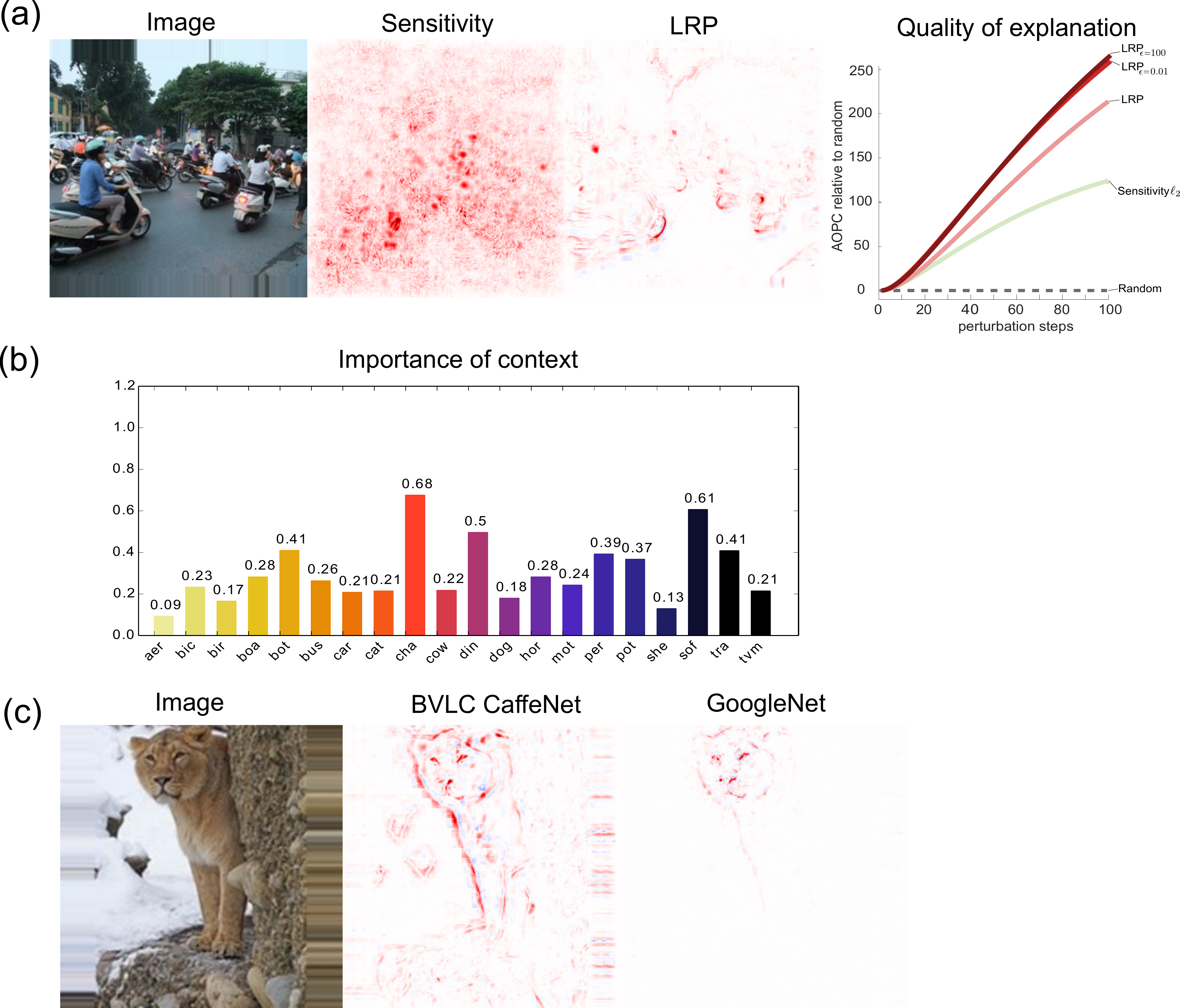} 
\caption{\label{fig:scooter}Experimental results. (a)~Qualitative and quantitative comparison of sensitivity analysis. (b)~Importance of context measured by LRP. (c)~Comparison of predictions by two DNN architectures. Illustrations are taken and adapted from the papers \cite{LapCVPR16} and \cite{binder-icml16}.}
\end{figure}

\paragraph{Comparing sensitivity analysis and LRP} A qualitative difference between the explanations produced by sensitivity analysis and LRP for the prediction of an image of class ``scooter'' is shown in Figure \ref{fig:scooter} (a). The explanations provided by sensitivity analysis are much noisier than the ones computed with LRP.
Regions consisting of pure background, e.g., the empty street, have large sensitivity, although these pixels are not really indicative for this image category.
However, if we put motor-bike like structures at these particular locations, then this change would certainly increase the classification score.
In contrast, the explanation provided by LRP does not highlight the locations where the classifier is very sensitive, but indicates how much every pixel contributes to the prediction.
Thus, it explains the given prediction and in this example only points to real scooter-like structures in the image. The explanations provided by LRP are not only better in a qualitative sense, but also quantitatively. The authors of \cite{SamTNNLS16} proposed an objective method for comparing visualizations based on perturbation analysis. The right plot in Figure \ref{fig:scooter} (a) displays the drop in classification score when perturbing regions identified by sensitivity analysis and LRP as being the most relevant ones (i.e., high $R_p$). A fast drop in classification score, i.e., a large AOPC value\footnote{AOPC stand for area over the perturbation curve.}, implies a meaningful sorting of $R_p$ and thus good explanations. The results in Figure \ref{fig:scooter} (a) clearly show that LRP provides better explanations than sensitivity analysis for images from the ILSVRC2012 dataset. More details about this experiment can be found in \cite{SamTNNLS16}.

\paragraph{Measuring importance of context in image classification} The explanations provided by LRP indicate how much every pixel contributes to the prediction. By aggregating the pixel-wise scores, e.g., inside and outside of an object bounding box, we can quantify the importance of context in image classification. Figure \ref{fig:scooter} (b) displays the outside-inside relevance ratio for the 20 categories of the PASCAL VOC 2007 dataset. For classes like ``airplane'' or ``sheep'' the context does not seem to be important for the DNN prediction as most relevance lies inside the bounding box.
This is different for indoor scene categories such as ``chair'' or ``sofa''. Here the LRP explanations show that context plays a much more important role for the classification (see \cite{LapCVPR16} for more discussion).

\paragraph{Comparing DNN architectures} Finally, we demonstrate the usage of LRP for analyzing differences between DNN architectures. Figure \ref{fig:scooter} (c) displays the LRP explanations obtained from BVLC CaffeNet \cite{DBLP:conf/mm/JiaSDKLGGD14} and GoogleNet \cite{DBLP:journals/corr/SzegedyLJSRAEVR14} when applied to an animal image. One can see that the latter network focuses more on the face of the animal and its explanation heatmap is significantly sparser than for BVLC CaffeNet. We observed this phenomenon for many animal images. This analysis suggests that GoogleNet found a better way to focus on the relevant information in the image, which is often the animal face and not the body shape or the fur.

\section{Conclusion}
The LRP method is a general framework for interpreting the predictions of complex ML systems such as deep neural networks, that can be used for model comparison, validation, or visualization. It was successfully applied to various tasks and machine learning models. Some redistribution rules for DNNs can be viewed as resulting from a deep Taylor decomposition of the neural network function~\cite{MonArXiv15}. Although the alpha-beta LRP rule was shown to work well for DNNs with ReLU activations, there is not one single LRP rule which serves all machine learning models and datasets equally well, as different models have specific layer-to-layer nonlinear mappings and input domains, that need to be considered.

\small

\bibliographystyle{IEEEtran}
\bibliography{nips2016}
\end{document}